\documentclass[letterpaper]{article} 
\usepackage{aaai2026}  
\usepackage{times}  
\usepackage{helvet}  
\usepackage{courier}  
\usepackage[hyphens]{url}  
\usepackage{graphicx} 
\usepackage{natbib}  
\usepackage{caption} 
\usepackage[ruled,vlined]{algorithm2e}
\usepackage{float}

\usepackage{newfloat}
\usepackage{listings}
\DeclareCaptionStyle{ruled}{labelfont=normalfont,labelsep=colon,strut=off} 
\floatstyle{ruled}
\newfloat{listing}{tb}{lst}{}
\floatname{listing}{Listing}
\title{
OIDA-QA: A Multimodal Benchmark for \\ Analyzing the Opioid Industry Documents Archive
}
\author{
    Xuan Shen\textsuperscript{\rm 1},
    Brian Wingenroth\textsuperscript{\rm 2}, 
    Zichao Wang\textsuperscript{\rm 3}, 
    Jason Kuen\textsuperscript{\rm 3}, 
    Wanrong Zhu\textsuperscript{\rm 3}, 
    Ruiyi Zhang\textsuperscript{\rm 3}, \\
    Yiwei Wang\textsuperscript{\rm 4}, 
    Lichun Ma\textsuperscript{\rm 5}, 
    Anqi Liu\textsuperscript{\rm 2}, 
    Hongfu Liu\textsuperscript{\rm 6}, 
    Tong Sun\textsuperscript{\rm 3},
    Kevin S. Hawkins\textsuperscript{\rm 2}, \\
    Kate Tasker\textsuperscript{\rm 7},
    G. Caleb Alexander\textsuperscript{\rm 2}, 
    Jiuxiang Gu\textsuperscript{\rm 3}\thanks{Corresponding author: Jiuxiang Gu (jigu@adobe.com)}
}
\affiliations{
    \textsuperscript{\rm 1}Northeastern University,
    \textsuperscript{\rm 2}Johns Hopkins University,
    \textsuperscript{\rm 3}Adobe Research
    \textsuperscript{\rm 4}University of California, Merced \\
    \textsuperscript{\rm 5}National Institutes of Health,
    \textsuperscript{\rm 6}Brandeis University,
    \textsuperscript{\rm 7}University of California, San Francisco 

    shen.xu@northeastern.edu, wingenroth@jhu.edu, \{jackwa, kuen, wzhu, ruizhang, tsun\}@adobe.com, \\
    yiweiwang2@ucmerced.edu, lichun.ma@nih.gov, aliu@cs.jhu.edu, hongfuliu@brandeis.edu, \\
    khawki29@jhu.edu, Kate.Tasker@ucsf.edu, galexan9@jhu.edu, jigu@adobe.com
}

\usepackage{bibentry}

\usepackage{enumitem}
\usepackage{comment}
\usepackage{amssymb}

\usepackage{graphicx}
\usepackage{amsmath} 
\usepackage{float}

\usepackage{multirow}
\usepackage{makecell}

\usepackage{microtype}
\usepackage{graphicx}
\usepackage{subfigure}
\usepackage{booktabs} 

\usepackage{caption}

\definecolor{darkblue}{rgb}{0.0, 0.0, 0.55}
\newcommand{\jx}[1]{\textcolor{black}{#1}}

\begin{document}

\maketitle

\begin{abstract}
The opioid crisis represents a significant moment in public health that reveals systemic shortcomings across regulatory systems, healthcare practices, corporate governance, and public policy. Analyzing how these interconnected systems simultaneously failed to protect public health requires innovative analytic approaches for exploring the vast amounts of data and documents disclosed in the UCSF-JHU Opioid Industry Documents Archive (OIDA).
The complexity, multimodal nature, and specialized characteristics of these healthcare-related legal and corporate documents necessitate more advanced methods and models tailored to specific data types and detailed annotations, ensuring the precision and professionalism in the analysis.
In this paper, we tackle this challenge by organizing the original dataset according to document attributes and constructing a benchmark with 400k training documents and 10k for testing. 
From each document, we extract rich multimodal information—including textual content, visual elements, and layout structures—to capture a comprehensive range of features. 
Using multiple AI models, we then generate a large-scale dataset comprising 360k training QA pairs and 10k testing QA pairs.
Building on this foundation, we develop domain-specific multimodal Large Language Models (LLMs) and explore the impact of multimodal inputs on task performance. To further enhance response accuracy, we incorporate historical QA pairs as contextual grounding for answering current queries.
Additionally, we incorporate page references within the answers and introduce an importance-based page classifier, further improving the precision and relevance of the information provided.
Preliminary results indicate the improvements with our AI assistant in document information extraction and question-answering tasks, 
highlighting the effectiveness of our benchmark.
The dataset and models are publicly available at \textcolor{blue}{\url{https://huggingface.co/datasets/opioidarchive/oida-qa}}.
\end{abstract}

\section{Introduction}

The opioid crisis has severely impacted global health, exposing weaknesses in healthcare systems and contributing to issues like domestic violence and child abuse~\citep{NIDA-Opioid, oderda2015economic, swedo2020adolescent}. In 2019, 10.1 million Americans reported opioid misuse, and from June 2021 to May 2022, opioids were involved in 90\% of the 108,000 U.S. overdose deaths~\citep{CDC-OPmisuse, CDC-OPdeath}. Though effective for pain relief, opioids can induce euphoria, resulting in misuse and addiction, particularly in regions with inadequate healthcare services, underscoring the urgent need to address drug misuse~\citep{birnbaum2011societal}.
Recent advances in AI have increasingly revealed its potential in healthcare.
Particularly, AI assistants based on \jx{LLMs} have emerged as powerful tools for extracting insights from unstructured medical data and assisting healthcare professionals in clinical Question Answering (QA)
~\citep{lee2020biobert, singhal2023large, Luclinicalbert}.
Thus, we wish to develop AI-driven solutions to address the opioid crisis with the public data from the UCSF-JHU Opioid Industry Documents Archive~\citep{OIDA}, a massive corpus of internal correspondence and other documents arising from the opioid industry.

While LLMs perform well on general QA tasks, the multimodal and long-context nature of healthcare data poses challenges such as complex reasoning and increased hallucination risks~\citep{gao2023retrieval}. 
Most LLMs also struggle with multi-turn interactions, limiting their ability to handle sequential user queries effectively. 
Furthermore, they often fail to link answers to specific pages or sections within documents, resulting in poor answer provenance and reduced trustworthiness.
These challenges are compounded by the escalating opioid crisis and the rapid growth of the OIDA dataset.
Thus, creating a low-cost, reliable, and scalable multimodal LLM accessible to the public is essential.


\begin{figure*}[t!]
\centering
\includegraphics[width=\textwidth]{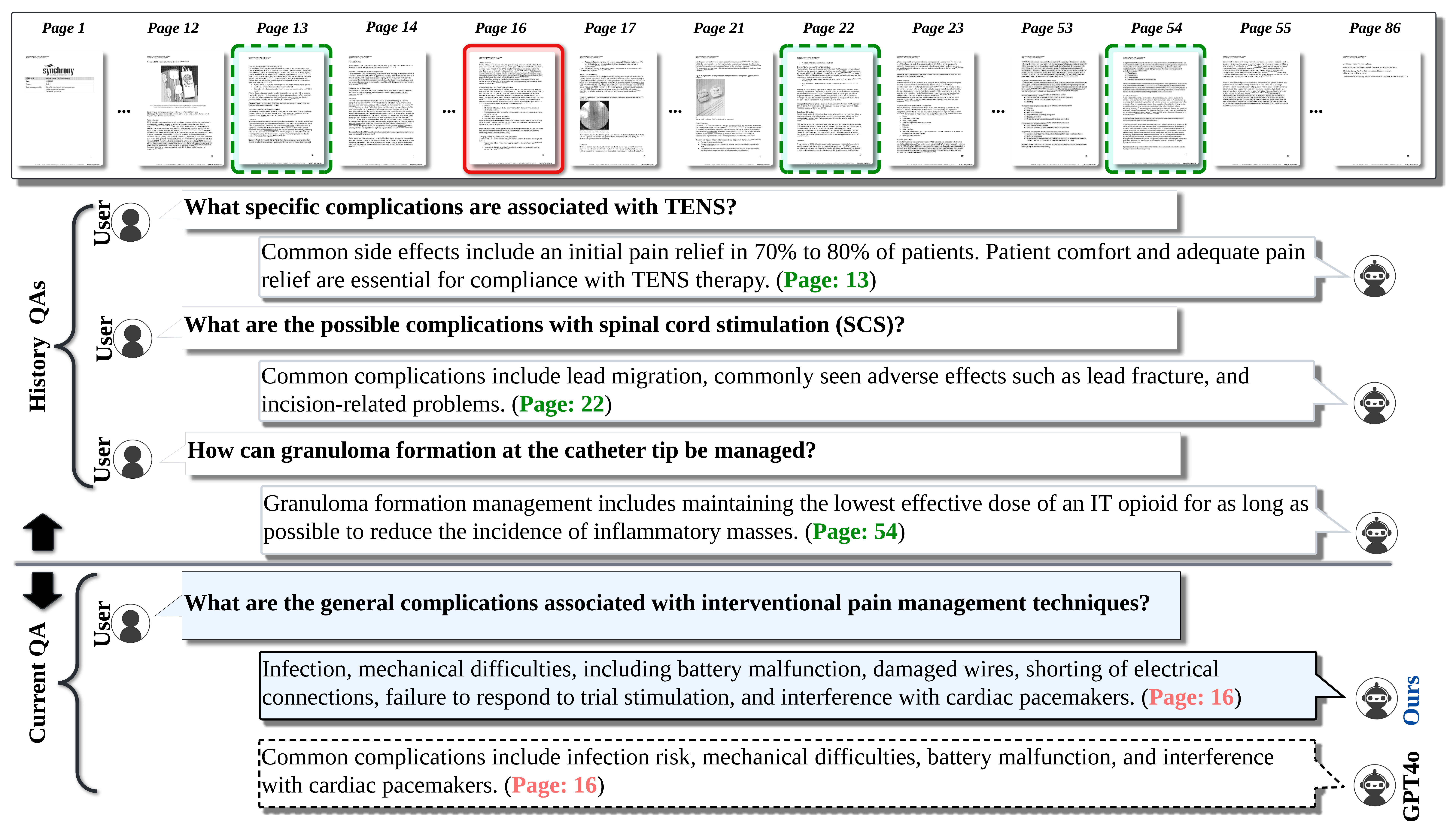}
\caption{Overview of the proposed OIDA-QA: Our QA dataset enables multi-hop QA with answers grounded to specific pages, effectively handling multi-page documents by extracting dense information (textual, visual, and layout) from scanned documents.}
\label{fig:oida_data_samples}
\end{figure*}

In this paper, we propose OIDA-QA, a multimodel multi-page and multi-hop document question-answering benchmark based on the OIDA.
To effectively handle large-scale document data, we start by analyzing its distribution using the taxonomy proposed in ADOPD~\citep{gu2024adopd}, combined with the CLIP~\citep{radford2021learning} model finetuned on ADOPD's image-caption pairs. Utilizing the taxonomy-derived clusters, we identify the 20 clusters with the most documents and further diversify based on sub-categories and page count. We compile 20K PDF documents from each selected cluster to create the final training set.
To develop a comprehensive understanding of each PDF document,
we enrich the original PDFs by extracting textual information (OCR words), visual elements (tags and masks), and layout details (bounding boxes). These enriched, model-assisted multimodal annotations provide the foundation for generating QA pairs.
Then, we introduce user personas~\citep{chan2024scaling} when generating questions, simulating users from various backgrounds to create a diverse set of queries.
Meanwhile, answerability and page-level grounding of the questions are ensured by leveraging a multimodal LLM, which serves as both a verifier and locator for the generated QA pairs.
Furthermore, \textbf{we recruit medical professionals, including doctors and nurses, to annotate and refine 100k QA pairs in our dataset, ensuring the accuracy and reliability of the majority of the generated QA pairs}.

To handle long-context reasoning, we enable LLMs to learn relations across extended contexts and ground answers to the right pages using instruction-based page-finding prompts and a key-page identification model.
By expanding the model's context window during training, we capture the semantic information of long documents. During testing, for documents that exceed the model's processing capacity, we use the key page identification model to preliminarily locate important pages and feed these into the model to obtain precise answers and their corresponding page locations. This approach effectively addresses the challenge of processing ultra-long texts, improving both efficiency and accuracy.
The overview of our proposed method is visualized in Figure~\ref{fig:oida_data_samples}.
Experiments show that our model performs excellently in both answering questions and locating pages, validating the effectiveness of our approach. This highlights the potential of multi-turn dialogue systems in the medical domain. We believe our system plays a vital role in medical information retrieval and opioid abuse prevention, providing robust technical support to address this pressing public health crisis.

The contributions of this paper are summarized as follows:
\setlength\intextsep{0pt}
\begin{itemize}[leftmargin=*, itemsep=1pt, topsep=0pt, partopsep=0pt, parsep=0pt]
    \item  We introduce OIDA-QA, a multimodal document QA benchmark based on the OIDA, along with an effective method for enriching PDF documents with textual, visual, and layout annotations. This provides a data foundation for the exploration of AI-driven solutions to the opioid crisis.
    \item  We utilize the dense extracted data and LLMs to create a persona-based, long-context, multi-page, multi-hop QA benchmark. By integrating user personas and page-level references, OIDA-QA benefits to a broader range of users and enhances the answer verifiability of the benchmark.
    \item We develop a scalable model system for long-context processing and answer page grounding. By employing instruction-based prompts and introducing a key page identification model, our approach enhances the model's ability to locate relevant pages within lengthy documents, improving both efficiency and accuracy.
    \item We validate the effectiveness of our proposed methods through extensive experiments. The results demonstrate strong performance in both question answering and page locating tasks, confirming the efficacy of our approach. 
\end{itemize}

\begin{figure*}[t!]
\centering
\includegraphics[width=1.0\textwidth]{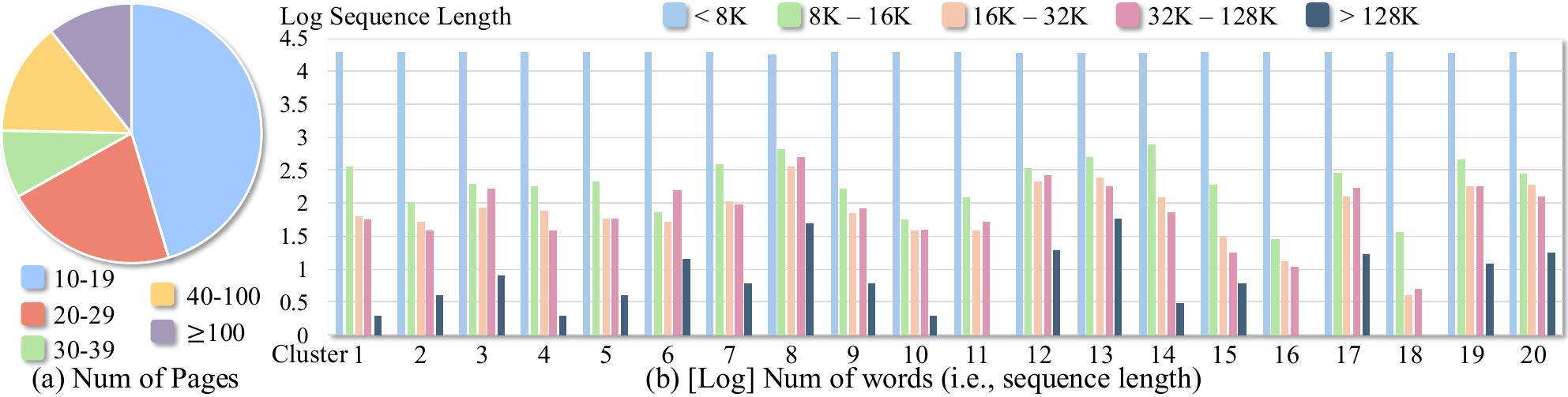}
\caption{Data distribution of (a) number of pages and (b) number of words (i.e., sequence length).}
\label{fig:oida_data_distribution}
\end{figure*}

\section{Related Work}
\paragraph{Document Understanding}

Pre-training large models that handle both textual and visual information have proven highly effective for document understanding tasks~\citep{huang2022layoutlmv3, da2023vision, shen2024hotaq, shen2024edgeqat, shen2024searchllm, shen2025draftattention, shen2025fastcar}. Unlike traditional LLMs that process plain text, document understanding requires models to consider {layout information}~\citep{tu-etal-2023-layoutmask, yu2023structextv, shen2025quartdepth, shen2024agile, shen2022vitlth, shen2025numerical, shen2025lazydit, shen2025sparse}.  Recently, LLMs and Multimodal LLMs (MLLMs)~\citep{chatgpt_webpage, openai2023gpt4, yang2023dawn, liu2025rora, liu2025toward} have demonstrated outstanding zero-shot performance across a wide range of Natural Language Processing (NLP) and Computer Vision (CV) tasks. Leveraging LLMs for zero-shot document understanding has also shown promise~\citep{perot2023lmdx, zhang2023llavar, shen2023deepmad, shen2025heima}. For example, LLaVAR~\citep{zhang2023llavar} extends LLaVA~\citep{liu2023llava, liu2023improvedllava} to the document domain by pre-training with OCR data, where the fine-tuning document instructions are generated by GPTs~\citep{achiam2023gpt}. 
Additionally, Qwen-VL~\citep{bai2023qwen} leverages document-level pre-training and direct QA for fine-tuning. Although existing models have demonstrated promising results in document tasks, most training data originates from traditional datasets that lack sufficient document data. 
The OIDA data poses challenges for existing models due to its multi-page, multimodal characteristics. Some works have been proposed for multi-image text generation tasks, such as InternVL~\citep{chen2023internvl, chen2024far} and~\cite{mistral2024pixtral}. However, these models were originally designed for general vision-language domain, not for multi-page document understanding tasks.

\paragraph{Healthcare Question Answering}
With over one-third of American adults seeking medical information online~\cite{Healthcare_report}, there is a growing need for robust healthcare QA systems to support patient consultations.
Previous healthcare QA systems with Pre-trained Language Models~\cite{pergola-etal-2021-boosting} faced challenges in real-world applications due to limited language understanding and generation capabilities~\cite{liu2023semantic}.
In healthcare, LLMs like Med-PaLM 2~\cite{singhal2023towards} have shown remarkable progress, achieving 86.5\% on the USMLE dataset and outperforming earlier models~\cite{singhal2022large}. 
This advancement extends to other medical datasets, including MedMCQA~\cite{pal2022medmcqa} and PubMedQA~\cite{jin2019pubmedqa}. 
However, as summarized in Table~\ref{tab:data_qa_comparsion}, existing healthcare QA datasets often lack multi-turn conversations and grounding information. 
Only datasets such as MedDialog-CN~\cite{he2020meddialog} and MedDialog-EN~\cite{he2020meddialog} provide multi-round dialogues but without grounding.
\begin{figure*}[t]
\centering
\includegraphics[width=1.0\textwidth]{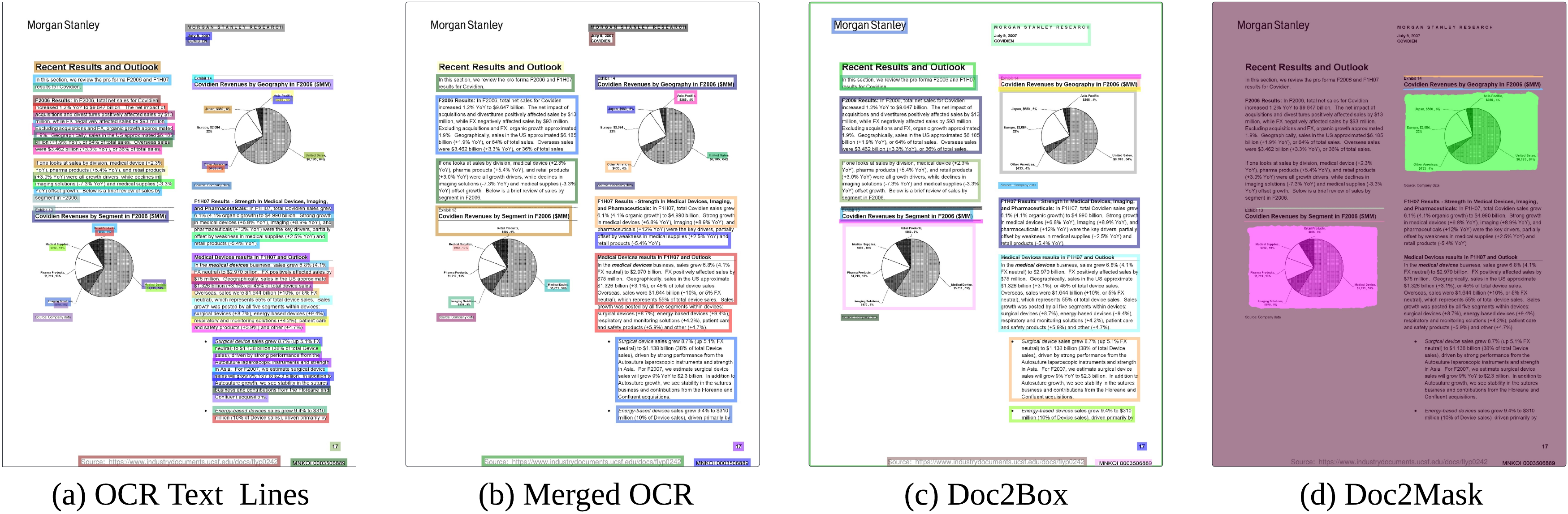}
\caption{Comparison of information extraction methods: (a) OCR text lines, (b) statistical merging results, (c) text blocks from Doc2Box, and (d) entity masks from Doc2Mask.}
\label{fig:oida_ocr_compare}
\end{figure*}

\section{OIDA-QA Benchmark}

\subsection{Data Collection and Extraction}\label{sec:data_extraction}

The original OIDA data is available in PDF format\footnote{\footnotesize{s3://opioid-industry-documents-archive-dataset-bucket/}}, along with metadata, and extracted text for all documents.
Layout and visual information are not available due to the limitations of OCR and document extraction models. For the OIDA documents, extracting as much detailed information as possible is essential to fully explore the model's capabilities and better serve the community. To fully leverage OIDA for developing an AI system, our data collection process includes the following steps: (1) analyzing the distribution of the OIDA dataset, (2) performing balanced data sampling, and (3) extracting multimodal information.

\noindent\textbf{Data Distribution Analysis}
To understand the distribution of OIDA, we utilize the pre-trained CLIP model from ADOPD and its taxonomy to tag the first page of each document in a zero-shot manner.
This is achieved by calculating the similarity between the visual features of the page images and the textual embeddings. The textual embeddings are composed of the proposed taxonomy labels from ADOPD, formatted as `\textit{a photo of <candidate label>}'.
The top five most relevant labels are selected for document grouping, which categorizes each document into clusters based on the hierarchically structured ADOPD taxonomy.
The predicted document tags and clustered results for all PDF documents offer a comprehensive overview of the OIDA dataset's distribution. This cluster-based sampling approach minimizes selection bias, resulting in a more diverse training set.

\noindent\textbf{Data Sampling}
Let the entire dataset be denoted by $\mathcal{D}_{\text{full}}$. We downsample $\mathcal{D}_{\text{full}}$ by selecting the top $K$ largest clusters based on distribution analysis above. For each cluster $k \in \{1, 2, \dots, K\}$, we define the subset $\mathcal{D}_k = \{D_{k,1}, D_{k,2}, \dots, D_{k,N_k}\}$, where each document $D_{k,i}$ consists of multiple pages. In our experiments, we set $K$ to 20. For the sampling of each cluster, we balance the subcategories according to the labels and the number of pages. We collect a diverse training set from these 20 clusters, including 20K PDF documents per cluster (for a total of 400K documents). For the test set, we gather 500 PDF documents per cluster, resulting in a total of 10K documents. 
To construct multi-hop QA pairs, we select long documents—those exceeding 10 pages—from the extracted dataset for better QA generation.
In Figure~\ref{fig:oida_data_distribution}(a), we visualize the distribution of documents across eight different page counts within each cluster. Figure~\ref{fig:oida_data_distribution}(b) visualizes the distribution of documents based on the logarithm of sequence length, emphasizing the long-sequence characteristics of OIDA-QA.

\noindent\textbf{Document Data Extraction}
For each document \( D_{k,i} \), we extract textual, visual, and layout information from each page. Using an OCR tool, we extract words to obtain a set of text lines, which are then grouped into paragraphs \( \{ \mathbf{p}_{k,i,j} \} \) using heuristic rules. We assign each paragraph \( \mathbf{p}_{k,i,j} \) a location \( \mathbf{l}_{k,i,j} = \left( p_{k,i,j}, b_x^l, b_y^t, b_x^r, b_y^b \right) \), where \( p_{k,i,j} \) is the page number, and \( (b_x^l, b_y^t, b_x^r, b_y^b) \) are the normalized bounding box coordinates. Figure~\ref{fig:oida_ocr_compare} illustrates that text lines alone do not capture the semantic relationships between words, and that merging words using rule-based methods is also limited by heuristic constraints. To address these limitations, we utilize the Doc2Box model~\citep{gu2024adopd} to extract text blocks that better preserve semantic structures. For visual information, we provide two outputs: {CLIP tags} and {entity masks}. The CLIP tags capture the high-level attributes of the documents. We apply the trained Doc2Mask model~\citep{gu2024adopd} to identify the entity masks. By combining textual, visual, and layout information, we create a comprehensive representation of each document that supports advanced processing tasks such as QA and information extraction.

\begin{figure*}[t]
\centering
\begin{minipage}{0.65\textwidth}
\centering
\vspace{-3mm}
\resizebox{\linewidth}{!}{
\begin{tabular}{llccc}
\toprule
\textbf{Task} & \textbf{Source} & \textbf{Size} & \textbf{\# Round} & \textbf{Grounding} \\
\midrule
\multirow{4}{*}{\rotatebox[origin=c]{90}{General}}
& Alpaca~\cite{alpaca} & 52k & 1 & No \\
& Dolly~\cite{dolly} & 15k & 1 & No \\
& ShareGPT~\cite{sharegpt} & 48k & 1 & No \\
& DocVQA~\cite{mathew2021docvqa} & 50k & 1 & No \\
\midrule
\multirow{8}{*}{\rotatebox[origin=c]{90}{HealthTech}}
& HealthCareMagic~\cite{wang2024_healthcaremagic100ken} & 100k & 1 & No \\
& iCliniq~\cite{icliniq} & 7k & 1 & No \\
& MedDialog-CN~\cite{he2020meddialog} & 792k & $\geq$1 & No \\
& MedDialog-EN~\cite{he2020meddialog} & 257k & $\geq$1 & No \\
& MedInstruct~\cite{medinstruct} & 52k & 1 & No \\
& Medical Flash Cards~\cite{medicalflashcards} & 34k & 1 & No \\
& WikiDocPatient~\cite{wikidocpatient} & 5k & 1 & No \\
& PubMedQA~\cite{jin2019pubmedqa} & 211k & 1 & No \\
\midrule
& \textbf{OIDA-QA (Ours)} & \textbf{370k} & \textbf{$\geq$2} & \textbf{Yes} \\
\bottomrule
\end{tabular}
}
\captionof{table}{
Comparison of QA datasets across general QA and HealthTech domains, including biomedical and clinical.
}
\label{tab:data_qa_comparsion}
\end{minipage}
\hfill
\begin{minipage}{0.33\textwidth}
    \centering
    \includegraphics[width=0.48\textwidth]{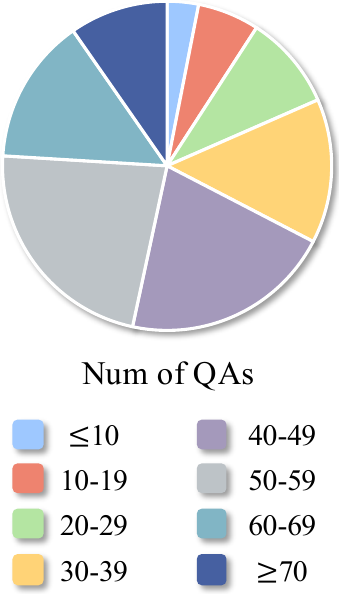}
    \captionof{figure}{Visualization of the QA pair counts within our benchmark.}
    \label{fig:qa_num_distribution}
\end{minipage}
\end{figure*}

\subsection{Persona-Based Multi-Hop Question-Answering Dataset Generation}

\noindent\textbf{Persona Setup}
To simulate diverse user interactions and generate questions from various perspectives, we incorporate the vast Persona Hub~\citep{chan2024scaling}, including over one billion personas, into our benchmark for question generation. We begin by generating the relevant personas for our benchmark.
For each cluster, we randomly sample 500 personas from the full Persona Hub and employ the GPT-4o~\citep{openai2023gpt4} to generate 48 detailed personas in average based on the assigned labels of the cluster.
The detailed personas, with attributes: \textit{Name}, \textit{Age}, \textit{Gender}, \textit{Major Background}, \textit{Previous Experience}, and \textit{Hobbies}, ensures the subsequent question generation process draws from a diverse user profiles.

\noindent\textbf{Multi-Hop QA Generation}
We employ a model-assisted approach for question-answer data generation, structuring the process into question generation and answer generation using GPT-4o.
Algorithm~\ref{alg:qa_generation_simple} detailed in Appendix illustrates the QA data generation process.
Building upon our data extraction method described in Section~\ref{sec:data_extraction}, we use only the grouped text lines as input for this step. To ensure a wide range of perspectives in the question generation process, we introduce the characterized personas discussed earlier. 
For generating QA pairs, we employ GPT-4o instances in the following role: (1) a question generator that creates questions based on the document content and persona attributes;
(2) an answer generator that determines if the answer is answerable and provides the corresponding response with referred page number simultaneously.
(3) a QA pair decomposer that decomposes single QA into more than one QA sequence to form a coherent multi-turn conversation.
As a result, we generate over 360k QA pairs with grounding information, indicating the specific page each answer refers to. Figure~\ref{fig:qa_num_distribution} shows the distribution of QA counts in our benchmark.

\section{Method}
\label{sec:method}

In this section, we detail the training process of our model for multi-round QA over long documents with page-grounded answers. Our approach includes instruction tuning of the LLM and introduces a page-finding mechanism, implemented by adjusting the training loss to improve page grounding and proposing a page-finding model to efficiently locate relevant pages within the documents.
  
\subsection{Instruction Tuning for Multi-Hop QA}\label{sec:multi_hop_qa}

We fine-tune the pretrained LLM using instruction tuning to perform multi-hop QA tasks. Given the LLM as model with parameters $\theta$, the model is trained to generate appropriate answers given an input context, a sequence of questions, and the dialogue history.  
  
Let $\mathcal{D} = \{(C_i, H_i)\}_{i=1}^{N}$ denote the training dataset, where $C_i$ is the input context for the $i$-th example, formulated based on the extracted sentences $\mathcal{S}$ and their locations; and $H_i = \left[ \left(q_i^k, a_i^k\right) \right]_{k=1}^{K_i}$ is the dialogue history consisting of question-answer pairs, with $K_i$ being the number of dialogue turns in the $i$-th example.
At each turn $j$, the model generates the answer $a_i^j$ conditioned on the context $C_i$, the current question $q_i^j$, and the previous dialogue history $H_i^{<j} = \left[ \left(q_i^k, a_i^k\right) \right]_{k=1}^{j-1}$; the conditional probability of generating $a_i^j$ is given by $P\left(a_i^j \mid C_i,\ q_i^j,\ H_i^{<j};\ \theta\right)$, where $\theta$ represents the model parameters. Finally, the training objective is to minimize the negative log-likelihood over the dataset as follows,
\begin{equation}  
    \mathcal{L}_{\text{QA}}(\theta) = -\sum_{i=1}^{N} \sum_{j=1}^{K_i} \log P\left(a_i^j \mid C_i,\ q_i^j,\ H_i^{<j};\ \theta\right).  
\end{equation}  
This loss function encourages the model to generate accurate answers based on the context and dialogue history in a multi-turn setting.  

\subsection{Query-Based Page Finding}
When dealing with long documents, the input context $C_i$ may exceed the model's maximum context window size $L_{\text{max}}$. To address this, we truncate the context to fit within the allowable length, typically by focusing on the ground-truth page. However, this introduces a training-testing mismatch, as the model relies on the ground-truth page during training but lacks this information during testing. To mitigate this, we propose two improvements: (1) \textit{Enhancing Page-Finding via Ground-Truth Content Reiteration} to improve content localization, and (2) \textit{Training a Query-based Page Finder} to enable effective page identification without relying on ground-truth information during testing.

\paragraph{Enhancing Page-Finding via Content Reiteration} 
The training objective in Section~\ref{sec:multi_hop_qa} focuses on optimizing the model's ability to generate accurate answers based on the input context and questions. To enhance the model's \textit{page-finding capability through content reiteration}, we introduce an additional optimization step. Specifically, we modify the model so that it outputs the associated page references along with a relevant portion of the context as a single, unified response. This is achieved by designing prompts that instruct the model to include page indices and pertinent context excerpts in its output, thereby reinforcing the link between the content and its location within the document.  
To formalize this enhancement, we adjust the conditional probability to jointly model the page indices and the relevant context portion as one output as follows, 
\begin{equation}  
P\left(p_i^j,\, \tilde{C}_i^j \mid C_i,\, q_i^j,\, H_i^{<j};\, \theta\right),  
\end{equation}  
where \( p_i^j \) represents the page indices relevant to the query, and \( \tilde{C}_i^j \) is the extracted portion of the context associated with those pages.  
  
Additionally, we further employ a page-finding loss function that captures the joint probability of generating the correct page and context to balance the learning of both tasks as follows,
\begin{equation}  
\mathcal{L}_{\text{PF}}(\theta) = -\sum_{i=1}^{N} \sum_{j=1}^{K_i} \log P\left(p_i^j,\, \tilde{C}_i^j \mid C_i,\, q_i^j,\, H_i^{<j};\, \theta\right).  
\end{equation}  
During training, we incorporate both the QA loss $\mathcal{L}_{\text{QA}}(\theta)$ and the page-finding loss $\mathcal{L}_{\text{PF}}(\theta)$ to enable accurate answer generation and effective page-level grounding over long documents.
We combine page-finding and QA data in a specific proportion to train the model on both tasks. This approach helps the model focus on relevant sections, output page references, and generate responses grounded in specific document parts, enhancing performance in multi-hop QA tasks.
  
\paragraph{Page Finder for Ultra-Long Contexts}  
As mentioned earlier, during testing, the input sequence can become very long, potentially exceeding hardware limitations and causing significant delays.
This motivates us to propose a method to assist in finding important pages during testing. Thus, we develop a separate \emph{Page Finder} module to identify relevant pages before generating answers.  
  
For multi-page documents, we first extract the content of each page. We use an encoder model based on Sentence Transformers~\citep{reimers2019sentence} to generate embeddings for each page \( c_l \) and the query \( q_i^j \) as follows,
\begin{equation}  
\mathbf{h}_l = \phi(c_l), \quad \mathbf{h}_q = \phi(q_i^j),  
\end{equation}  
where \( \phi \) denotes the Sentence Transformer encoder.  
  
To train the Page Finder module, we utilize the Multiple Negatives Ranking Loss~\citep{henderson2017efficient}, which is effective for training retrieval models with in-batch negatives. Given a batch of \( B \) query-document pairs \( \{ (q_b, c_b^+) \}_{b=1}^{B} \), where each \( q_b \) is a query and \( c_b^+ \) is its corresponding positive (relevant) page, we compute the similarity scores between each query and all document embeddings in the batch. Then, the total loss over the batch is then computed as follows,  
\begin{equation}  
\mathcal{L}_{\text{MNRL}}=- \frac{1}{B} \sum_{b=1}^{B} \log \frac{\exp(s_{b,b} / \tau)}{\sum_{k=1}^{B} \exp(s_{b,k} / \tau)},  
\end{equation}  
where \( s_{b,k} = \cos(\mathbf{h}_{q_b},\, \mathbf{h}_{c_k^+}) \) is the similarity score between query \( q_b \) and document \( c_k^+ \), and \( \tau \) is a temperature hyperparameter that controls the softness of the probability distribution. This loss encourages the model to assign high similarity scores to the correct query-document pairs while minimizing the scores for mismatched pairs within the batch, which serve as negative samples. 
  
During inference, we employ the trained Page Finder to compute the relevance scores between the query and each page using cosine similarity. We then select the top-ranked pages based on these scores and extend the selection by adding additional adjacent pages until the context length limit \( L_{\text{max}} \) is reached as follows,  
\begin{equation}  
C_i^{\text{reduced}} = [ c_{l_1},\ c_{l_2},\ \dots,\ c_{l_K} ],  
\end{equation}  
where \( K \) is determined such that the combined length of the selected pages does not exceed \( L_{\text{max}} \).  
  
We then provide \( C_i^{\text{reduced}} \) to LLMs for answer generation as follows,
\begin{equation}  
P\left( a_i^j \mid C_i^{\text{reduced}},\ q_i^j,\ H_i^{<j};\ \theta \right).  
\end{equation}
By integrating the Page Finder module into our system, we effectively handle ultra-long documents by focusing on the most relevant sections, mitigating hardware limitations and reducing inference delay. Meanwhile, training the Page Finder enables it to accurately retrieve the pages most pertinent to the query, enhancing the overall performance.

\section{Experiments}

\subsection{Experimental Analysis}

\paragraph{Training Details}
In our main experiments, we utilize the Mistral-7B-Instruct-v0.2 model~\cite{jiang2023mistral}, employing full-parameter fine-tuning for improved optimization. The training is conducted over one epoch with a local batch size of 12 and a learning rate set to \(5 \times 10^{-6}\), utilizing 8 NVIDIA H100 GPUs.  
The AdamW optimizer~\cite{loshchilov2018decoupled} is used with cross-entropy loss, and the maximum sequence length is set to 8192.
Additionally, we derive an extra 64,000 QA training examples from the original training set, specifically tailored for content reiteration. \jx{For the Page Finder module, we utilize the \texttt{multi-qa-mpnet-base-dot-v1} model~\citep{reimers2021sentence}, and we further fine-tune it on 100{,}000 random samples generated from the training set.}
The fine-tuning process involves training for one epoch on 8 NVIDIA H100 GPUs, with a local batch size of 16, a learning rate set to \(2 \times 10^{-5}\), and a warmup ratio of 0.1.

\paragraph{Evaluation Metrics}
We assess the generated answers using sentence-level automatic evaluation metrics provided by the Hugging Face evaluation pipeline~\citep{hg_eval_pipeline}. Automated metrics include BLEU~\citep{papineni2002bleu}, METEOR~\citep{banerjee2005meteor}, ROUGE~\citep{lin2004rouge}, and BERTScore~\citep{zhang2019bertscore}, which provide quantitative measures of answer quality by comparing generated responses to reference answers.
We also introduce the \textit{page generation rate} and \textit{page accuracy} to further demonstrate the effectiveness of our proposed page-finding optimization method. The \textit{page generation rate} denotes the proportion of generated answers that include a reference to a page. The \textit{page accuracy} refers to the proportion of these generated page references that correctly correspond to the intended reference pages.

\begin{table*}[t!]
\centering
\resizebox{1.0\linewidth}{!}{
\begin{tabular}{c|c|c|cccc|c|cccc|c}
\toprule
\textbf{Window} & \textbf{Content} & \textbf{Page} & \multicolumn{4}{c|}{\textbf{BLEU}} & \multirow{2}{*}{\textbf{METEOR}} & \multicolumn{4}{c|}{\textbf{ROUGE}} & \textbf{BERT} \\  
\cline{4-7}\cline{9-12}
\textbf{Size} & \textbf{Reiteration} & \textbf{Finder} & \textbf{-1} & \textbf{-2} & \textbf{-3} & \textbf{-4} &  & \textbf{-1} & \textbf{-2} & \textbf{-L} & \textbf{Lsum} & \textbf{Score} \\
\midrule
/  & $\times$ & $\times$          & 65.9\%                  & 52.4\%                  & 45.2\%                  & 38.5\%                  & 60.9\%                  & 56.8\%                   & 42.2\%                   & 53.6\%                   & 53.7\% & 88.5\% \\
/  & $\times$ & $\checkmark$                      & 73.5\%                  & 63.7\%                  & 57.2\%                  & 48.6\%                  & 67.9\%                  & 63.9\%                   & 51.9\%                   & 61.7\%                   & 61.7\% & 90.7\% \\
\midrule
1      & $\times$ & $\times$          & 74.6\%                  & 62.8\%                  & 55.9\%                  & 49.7\%                  & 69.5\%                  & 66.1\%                   & 53.5\%                   & 63.7\%                   & 63.7\% & 91.7\% \\
1      & $\times$ & $\checkmark$                      & 73.6\%                  & 61.5\%                  & 54.4\%                  & 48.1\%                  & 67.8\%                  & 64.1\%                   & 51.1\%                   & 61.7\%                   & 61.7\% & 91.3\% \\
1      & $\checkmark$             & $\times$          & 75.6\%                  & 64.1\%                  & 57.1\%                  & 51.0\%                  & 69.4\%                  & 66.5\%                   & 54.1\%                   & 64.3\%                   & 64.3\% & 91.8\% \\
1      & $\checkmark$             & $\checkmark$                      & 77.0\%                  & 66.3\%                  & 59.6\%                  & 53.7\%                  & 71.7\%                  & 68.9\%                   & 56.8\%                   & 66.5\%                   & 66.5\% & 92.3\% \\
\midrule
3      & $\times$ & $\times$          & 71.6\%                  & 61.1\%                  & 54.8\%                  & 48.7\%                  & 67.7\%                  & 63.5\%                   & 51.9\%                   & 61.2\%                   & 61.3\% & 90.6\% \\
3      & $\times$ & $\checkmark$                      & 73.3\%                  & 63.4\%                  & 57.3\%                  & 52.0\%                  & 70.5\%                  & 66.9\%                   & 55.5\%                   & 64.7\%                   & 64.7\% & 91.6\% \\
3      & $\checkmark$             & $\times$          & 72.6\%                  & 62.0\%                  & 55.4\%                  & 49.5\%                  & 68.9\%                  & 64.9\%                   & 52.9\%                   & 62.6\%                   & 62.6\% & 91.1\% \\
3      & $\checkmark$             & $\checkmark$                      & 75.9\%                  & 65.9\%                  & 59.4\%                  & 53.6\%                  & 71.4\%                  & 68.1\%                   & 56.4\%                   & 65.8\%                   & 65.8\% & 91.8\% \\
\midrule
Max    & $\times$ & $\times$          & 72.3\%                  & 61.8\%                  & 55.3\%                  & 49.8\%                  & 70.6\%                  & 66.2\%                   & 54.1\%                   & 63.8\%                   & 63.8\% & 91.5\% \\
Max    & $\times$ & $\checkmark$                      & 73.6\%                  & 63.8\%                  & 57.6\%                  & 52.2\%                  & 72.4\%                  & 68.0\%                   & 56.4\%                   & 65.6\%                   & 65.6\% & 91.9\% \\
Max    & $\checkmark$             & $\times$          & 75.1\%                  & 64.1\%                  & 57.4\%                  & 51.4\%                  & 69.7\%                  & 66.7\%                   & 54.4\%                   & 64.4\%                   & 64.4\% & 91.7\% \\
Max    & $\checkmark$             & $\checkmark$                      & 76.5\%                  & 66.2\%                  & 59.7\%                  & 54.0\%                  & 71.7\%                  & 68.7\%                   & 56.8\%                   & 66.4\%                   & 66.4\% & 92.2\% \\
\bottomrule
\end{tabular}
}
\caption{
Main results with different settings, including varying window sizes and the use of content reiteration or the \textit{Page Finder} module. Window size refers to the number of adjacent pages included before and after the ground-truth page. The maximum window size denotes truncation centered around the ground-truth page, extending to the maximum sequence length allowed.
}
\label{tab:results_main}
\end{table*}

\subsection{Experimental Analysis}  
\label{sec:experimental_analysis}

\paragraph{Impact of Context Window Size on Model Performance}

\begin{table}[t!]
\centering
\resizebox{1.0\linewidth}{!}{
\begin{tabular}{c|c|c|cc}
\toprule
\textbf{Window} & \textbf{Page}    & \textbf{Page}   & \textbf{Page}        & \textbf{Page}   \\
      \textbf{Size}       & \textbf{Finding} & \textbf{Finder} & \textbf{Generation Rate} & \textbf{Accuracy} \\
                        \midrule
/  & $\times$ & $\times$ & 68.7\%   & 83.2\%          \\
/  & $\times$ & $\checkmark$             & 88.2\%   & 90.8\%          \\
\midrule
1      & $\times$ & $\times$ & 69.7\%   & 98.9\%          \\
1      & $\times$ & $\checkmark$             & 71.0\%   & 99.1\%          \\
1      & $\checkmark$             & $\times$ & 73.9\%   & 99.1\%          \\
1      & $\checkmark$             & $\checkmark$             & 83.4\%   & 99.2\%          \\
\midrule
3      & $\times$ & $\times$ & 80.4\%   & 97.8\%          \\
3      & $\times$ & $\checkmark$             & 87.5\%   & 98.5\%          \\
3      & $\checkmark$             & $\times$ & 79.2\%   & 96.6\%          \\
3      & $\checkmark$             & $\checkmark$             & 88.1\%   & 98.0\%          \\
\midrule
Max    & $\times$ & $\times$ & 80.8\%   & 98.8\%          \\
Max    & $\times$ & $\checkmark$             & 86.7\%   & 99.2\%          \\
Max    & $\checkmark$             & $\times$ & 80.8\%   & 97.6\%          \\
Max    & $\checkmark$             & $\checkmark$             & 88.5\%   & 97.6\%           \\
\bottomrule
\end{tabular}
}
\caption{
Page generation rate and page accuracy results. The generation rate represents the proportion of generated answers that include the reference page.
}
\label{tab:results_page_acc}
\end{table}

In our experiments with long documents, we investigate the effect of various context window sizes on model performance. Specifically, we use context window sizes of 0, 1, 3, and the maximum allowable size. 
Maximum window size includes as many adjacent pages as possible until reaching the maximum sequence length permitted by our experimental setup.  
  
Results in Table~\ref{tab:results_main} highlight the critical importance of context window size. Training without any context window (window size of 0) yields poorest performance, as conditioning on inputs without relevant context leads to unstable loss optimization. 
As context window size increases, the model's performance improves significantly, particularly in terms of \textit{page accuracy} and \textit{page generation rate}, as shown in Table~\ref{tab:results_page_acc}. 
Incorporating a larger context window enhances the model's ability to accurately identify and generate relevant pages. 
These findings highlight the importance of using adjacent page information to improve model performance on multi-page document tasks, while balancing hardware constraints from memory and bandwidth for optimal results.

\paragraph{Enhancing Model Performance with Content Reiteration} 

To further strengthen the model's page-finding capabilities, we implement a content reiteration strategy. This involves randomly sampling training examples and reformulating the QA pairs into a page-finding format. These reformulated samples are then incorporated into the original training set, enabling optimized instruction tuning.
The experimental results in Table~\ref{tab:results_main} indicate that content reiteration-guided instruction tuning significantly enhances the model's reading comprehension and page localization abilities. This approach proves particularly effective when the context window size is small, as presented in Table~\ref{tab:results_page_acc}, by reinforcing the model's ability to pinpoint relevant pages.  
However, content reiteration alone may not fully compensate for the limitations of a small window size. Despite its benefits, the page accuracy for models trained with a small window size remains lower than that of models trained with the maximum window size, even without the use of content reiteration. This highlights that larger context windows are still crucial for optimal performance.  
  

\paragraph{Impact of the Page Finder on Model Performance}  
  
We further enhance our benchmark by incorporating the Page Finder module and conducting additional evaluations to verify its effectiveness, as shown in Table~\ref{tab:results_main} and Table~\ref{tab:results_page_acc}. The inclusion of the Page Finder allows us to mitigate practical limitations associated with processing very long documents, especially on mobile devices with constrained GPU memory. Inputting excessively long documents is often impractical due to hardware limitations and can introduce irrelevant information that may hinder the model's performance.  

  
Also, this groundwork paves the way for efficient deployment in real-world applications, enabling scalable solutions with rapid response for complex medical information retrieval
in healthcare and public health contexts.

We visualize the comparison to GPT-4 in Figure~\ref{fig:oida_vis_samples} at Appendix. 
The results demonstrate   
our model's ability to perform multi-hop question answering by effectively utilizing contextual information across multiple pages. It not only provides accurate, contextually relevant responses but also precisely identifies the corresponding pages in the document.

\section{Conclusion and Future Work}

This paper introduces OIDA-QA, a multimodal QA benchmark designed specifically for OIDA document understanding. OIDA-QA comprises 400K training documents and 10K testing documents. In addition, we have collected over 370k multi-hop QA pairs generated by various models to enhance the dataset's diversity and robustness. Our experimental results demonstrate the effectiveness of both the benchmark and the AI assistant system in tackling multi-round QA tasks, presenting a promising approach to addressing opioid crisis.


\section*{Author Disclosure}

Dr. Alexander is past Chair of FDA’s Peripheral and Central Nervous System Advisory Committee and a co-founding Principal and equity holder in Stage Analytics.  These arrangements have been reviewed and approved by Johns Hopkins University in accordance with its conflict-of-interest policies.
\bibliography{reference}

@misc{reimers2021sentence,
  title         = {Sentence-Transformers: Multilingual Sentence, Paragraph, and Image Embeddings using BERT \& Co.},
  author        = {Nils Reimers and Iryna Gurevych},
  year          = {2021},
  howpublished  = {\url{https://huggingface.co/sentence-transformers/multi-qa-mpnet-base-dot-v1}},
  note          = {Accessed: 2025}
}

@article{chan2024scaling,
  title={Scaling synthetic data creation with 1,000,000,000 personas},
  author={Chan, Xin and Wang, Xiaoyang and Yu, Dian and Mi, Haitao and Yu, Dong},
  journal={arXiv preprint arXiv:2406.20094},
  year={2024}
}

@misc{mistral2024pixtral,
  author       = {Pixtral},
  title        = {Pixtral 12B: Unleashing the Power of 12 Billion Parameters},
  url          = {https://mistral.ai/news/pixtral-12b/},
  note         = {Accessed: [Date]},
year=2024
}

@article{chen2023internvl,
      title={InternVL: Scaling up Vision Foundation Models and Aligning for Generic Visual-Linguistic Tasks},
      author={Chen, Zhe and Wu, Jiannan and Wang, Wenhai and Su, Weijie and Chen, Guo and Xing, Sen and Zhong, Muyan and Zhang, Qinglong and Zhu, Xizhou and Lu, Lewei and Li, Bin and Luo, Ping and Lu, Tong and Qiao, Yu and Dai, Jifeng},
      journal={arXiv preprint arXiv:2312.14238},
      year={2023}
  }

@article{chen2024far,
    title={How Far Are We to GPT-4V? Closing the Gap to Commercial Multimodal Models with Open-Source Suites},
    author={Chen, Zhe and Wang, Weiyun and Tian, Hao and Ye, Shenglong and Gao, Zhangwei and Cui, Erfei and Tong, Wenwen and Hu, Kongzhi and Luo, Jiapeng and Ma, Zheng and others},
    journal={arXiv preprint arXiv:2404.16821},
    year={2024}
  }

@Manual{NIDA-Opioid,
	title ={Opioids},
	author ={NIDA},
	year ={2024},
	note ={\url{https://www.drugabuse.gov/drug-topics/opioids}}
}

@article{singhal2023large,
  title={Large language models encode clinical knowledge},
  author={Singhal, Karan and Azizi, Shekoofeh and Tu, Tao and Mahdavi, S Sara and Wei, Jason and Chung, Hyung Won and Scales, Nathan and Tanwani, Ajay and Cole-Lewis, Heather and Pfohl, Stephen and others},
  journal={Nature},
  volume={620},
  number={7972},
  pages={172--180},
  year={2023},
  publisher={Nature Publishing Group}
}

@Manual{CDC-OPmisuse,
	title ={Americans Share Hopeful Stories of Recovery From Opioid Use Disorder},
	author ={CDC},
	year ={2020},
	note ={\url{https://www.cdc.gov/rxawareness/pdf/articles/TA-T3D2-English_MatteArticle_Release_508.pdf}}
}

@article{birnbaum2011societal,
  title={Societal costs of prescription opioid abuse, dependence, and misuse in the United States},
  author={Birnbaum, Howard G and White, Alan G and Schiller, Matt and Waldman, Tracy and Cleveland, Jody M and Roland, Carl L},
  journal={Pain medicine},
  volume={12},
  number={4},
  pages={657--667},
  year={2011},
  publisher={Blackwell Publishing Inc Malden, USA}
}

@article{CDC-OPdeath,
	title={Drug Overdose Deaths Among Persons Aged 10–19 Years - United States, July 2019-December 2021},
	author={Tanz, Lauren J.  and Dinwiddie, Amanda T. and Mattson, Christine L.  and O’Donnell, Julie  and Davis, Nicole L.},
	journal={Morbidity and Mortality Weekly Report},
	year={2022},
	publisher={CDC}
}

@inproceedings{mathew2021docvqa,
  title={Docvqa: A dataset for vqa on document images},
  author={Mathew, Minesh and Karatzas, Dimosthenis and Jawahar, CV},
  booktitle={WACV},
  pages={2200--2209},
  year={2021}
}

@inproceedings{huang2022layoutlmv3,
  title={LayoutLMv3: Pre-training for Document AI with Unified Text and Image Masking},
  author={Huang, Yupan and Lv, Tengchao and Cui, Lei and Lu, Yutong and Wei, Furu},
  booktitle={ACM Multimedia},
  year={2022}
}

@misc{OIDA,
  author       = {OIDA},
  title        = {Opioid Industry Documents Archive},
  howpublished = {\url{https://www.industrydocuments.ucsf.edu/opioids/}},
  note         = {Accessed: 2024-10-01},
year=2021
}

@inproceedings{yu2023structextv,
    title={StrucTexTv2: Masked Visual-Textual Prediction for Document Image Pre-training},
    author={Yuechen Yu and Yulin Li and Chengquan Zhang and Xiaoqiang Zhang and Zengyuan Guo and Xiameng Qin and Kun Yao and Junyu Han and Errui Ding and Jingdong Wang},
    booktitle={ICLR},
    year={2023}
}

@inproceedings{da2023vision,
  title={Vision Grid Transformer for Document Layout Analysis},
  author={Da, Cheng and Luo, Chuwei and Zheng, Qi and Yao, Cong},
  booktitle={ICCV},
  year={2023}
}

@article{liu2023improvedllava,
  title={Improved Baselines with Visual Instruction Tuning},
  author={Liu, Haotian and Li, Chunyuan and Li, Yuheng and Lee, Yong Jae},
  journal={arXiv preprint arXiv:2310.03744},
  year={2023}
}

@article{zhang2023llavar,
  title={Llavar: Enhanced visual instruction tuning for text-rich image understanding},
  author={Zhang, Yanzhe and Zhang, Ruiyi and Gu, Jiuxiang and Zhou, Yufan and Lipka, Nedim and Yang, Diyi and Sun, Tong},
  journal={arXiv preprint arXiv:2306.17107},
  year={2023}
}

@article{bai2023qwen,
  title={Qwen-vl: A frontier large vision-language model with versatile abilities},
  author={Bai, Jinze and Bai, Shuai and Yang, Shusheng and Wang, Shijie and Tan, Sinan and Wang, Peng and Lin, Junyang and Zhou, Chang and Zhou, Jingren},
  journal={arXiv preprint arXiv:2308.12966},
  year={2023}
}

@article{perot2023lmdx,
  title={LMDX: Language Model-based Document Information Extraction and Localization},
  author={Perot, Vincent and Kang, Kai and Luisier, Florian and Su, Guolong and Sun, Xiaoyu and Boppana, Ramya Sree and Wang, Zilong and Mu, Jiaqi and Zhang, Hao and Hua, Nan},
  journal={arXiv preprint arXiv:2309.10952},
  year={2023}
}

@misc{chatgpt_webpage,
  title={Introducing chatgpt},
  author={OpenAI},
  howpublished = {\url{https://openai.com/blog/chatgpt}},
  year={2022}
}

@inproceedings{radford2021learning,
  title={Learning transferable visual models from natural language supervision},
  author={Radford, Alec and Kim, Jong Wook and Hallacy, Chris and Ramesh, Aditya and Goh, Gabriel and Agarwal, Sandhini and Sastry, Girish and Askell, Amanda and Mishkin, Pamela and Clark, Jack and others},
  booktitle={ICML},
  year={2021}
}

@misc{alpaca,
  author = {Rohan Taori and Ishaan Gulrajani and Tianyi Zhang and Yann Dubois and Xuechen Li and Carlos Guestrin and Percy Liang and Tatsunori B. Hashimoto },
  title = {Stanford Alpaca: An Instruction-following LLaMA model},
  year = {2023},
  publisher = {GitHub},
  journal = {GitHub repository},
  howpublished = {\url{https://github.com/tatsu-lab/stanford_alpaca}},
}

@inproceedings{tu-etal-2023-layoutmask,
    title = "{L}ayout{M}ask: Enhance Text-Layout Interaction in Multi-modal Pre-training for Document Understanding",
    author = "Tu, Yi  and
      Guo, Ya  and
      Chen, Huan  and
      Tang, Jinyang",
    booktitle = "ACL",
    year = "2023",
}

@article{openai2023gpt4,
  title={GPT-4 technical report},
  author={OpenAI, R},
  journal={arXiv},
  pages={2303--08774},
  year={2023}
}

@article{yang2023dawn,
  title={The dawn of lmms: Preliminary explorations with {GPT-4V(ision)}},
  author={Yang, Zhengyuan and Li, Linjie and Lin, Kevin and Wang, Jianfeng and Lin, Chung-Ching and Liu, Zicheng and Wang, Lijuan},
  journal={arXiv preprint arXiv:2309.17421},
  volume={9},
  year={2023}
}

@misc{liu2023llava,
      title={Visual Instruction Tuning}, 
      author={Liu, Haotian and Li, Chunyuan and Wu, Qingyang and Lee, Yong Jae},
      publisher={arXiv:2304.08485},
      year={2023},
}

@inproceedings{papineni2002bleu,
  title={Bleu: a method for automatic evaluation of machine translation},
  author={Papineni, Kishore and Roukos, Salim and Ward, Todd and Zhu, Wei-Jing},
  booktitle={ACL},
  year={2002}
}

@article{achiam2023gpt,
  title={Gpt-4 technical report},
  author={Achiam, Josh and Adler, Steven and Agarwal, Sandhini and Ahmad, Lama and Akkaya, Ilge and Aleman, Florencia Leoni and Almeida, Diogo and Altenschmidt, Janko and Altman, Sam and Anadkat, Shyamal and others},
  journal={arXiv preprint arXiv:2303.08774},
  year={2023}
}

@misc{jiang2023mistral,
      title={Mistral 7B}, 
      author={Albert Q. Jiang and Alexandre Sablayrolles and Arthur Mensch and Chris Bamford and Devendra Singh Chaplot and Diego de las Casas and Florian Bressand and Gianna Lengyel and Guillaume Lample and Lucile Saulnier and Lélio Renard Lavaud and Marie-Anne Lachaux and Pierre Stock and Teven Le Scao and Thibaut Lavril and Thomas Wang and Timothée Lacroix and William El Sayed},
      year={2023},
      eprint={2310.06825},
      archivePrefix={arXiv},
      primaryClass={cs.CL}
}

@misc{dolly,  
  title = {Databricks Dolly: Documentation and Resources},  
  author = {Databricks},  
  year = {2023},  
  howpublished = {\url{https://www.databricks.com/blog/2023/03/24/hello-dolly-democratizing-the-magic-of-chatgpt-with-open-models.html}}  
}

@misc{sharegpt,  
  title = {ShareGPT},  
  author = {ShareGPT},  
  year = {2023},  
  howpublished = {\url{https://huggingface.co/datasets/anon8231489123/ShareGPT_Vicuna_unfiltered}}  
}

@misc{icliniq,  
  title = {iCliniq Dataset},  
  author = {iCliniq},  
  note = {\url{https://www.icliniq.com/}}  
}

@misc{medinstruct,  
  title = {MedInstruct: A Large-scale Medical Instruction Dataset},  
  author = {Gu, Yu and Tinn, Robert and Cheng, Hao and et al.},  
  year = {2023},  
  howpublished = {\url{https://github.com/microsoft/medinstruct}}  
}

@misc{medicalflashcards,  
  title = {Medical Flashcards Dataset},  
  author = {OpenAnesthesia},  
  note = {\url{https://www.openanesthesia.org/keywords/}}  
}

@misc{wikidocpatient,  
  title = {WikiDocPatient Dataset},  
  author = {Curai Research},  
  year = {2020},  
  howpublished = {\url{https://github.com/curai/curai-research/tree/master/WikiDoc-Patient}}  
}

@inproceedings{lin2004rouge,  
  title={ROUGE: A package for automatic evaluation of summaries},  
  author={Lin, Chin-Yew},  
  booktitle={Text Summarization Branches Out},  
  pages={74--81},  
  year={2004}  
}

@misc{hg_eval_pipeline,
      title={A library for easily evaluating machine learning models and datasets.},
      author={HuggingFace},
      year={2024},
    url={https://huggingface.co/docs/evaluate/index}
}

@article{zhang2019bertscore,  
  title={BERTScore: Evaluating text generation with BERT},  
  author={Zhang, Tianyi and Kishore, Varsha and Wu, Felix and Weinberger, Kilian Q and Artzi, Yoav},  
  journal={arXiv preprint arXiv:1904.09675},  
  year={2019}  
}

@article{loshchilov2018decoupled,  
  title={Decoupled Weight Decay Regularization},  
  author={Loshchilov, Ilya and Hutter, Frank},  
  journal={ICLR},  
  year={2018}  
}

@inproceedings{henderson2017efficient,  
  title={Efficient Natural Language Response Suggestion for Smart Reply},  
  author={Henderson, Matthew and Iyer, Srinivasan and Golub, David and Agarwal, Sagar and Guo, Deming and Huang, Ji and Mayo, Douglas and Chen, Rumeng and Tseng, Belle and Siminian, Maryam and others},  
  booktitle={Proceedings of the 23rd ACM SIGKDD International Conference on Knowledge Discovery and Data Mining},  
  pages={331--339},  
  year={2017}  
}

@inproceedings{reimers2019sentence,  
  title={Sentence-BERT: Sentence Embeddings using Siamese BERT-Networks},  
  author={Reimers, Nils and Gurevych, Iryna},  
  booktitle={Proceedings of the 2019 Conference on Empirical Methods in Natural Language Processing},  
  pages={3982--3992},  
  year={2019}  
}

@article{gao2023retrieval,
  title={Retrieval-augmented generation for large language models: A survey},
  author={Gao, Yunfan and Xiong, Yun and Gao, Xinyu and Jia, Kangxiang and Pan, Jinliu and Bi, Yuxi and Dai, Yi and Sun, Jiawei and Wang, Haofen},
  journal={arXiv preprint arXiv:2312.10997},
  year={2023}
}

@inproceedings{
gu2024adopd,
title={{ADOPD}: A Large-Scale Document Page Decomposition Dataset},
author={Jiuxiang Gu and Xiangxi Shi and Jason Kuen and Lu Qi and Ruiyi Zhang and Anqi Liu and Ani Nenkova and Tong Sun},
booktitle={ICLR},
year={2024}
}

@article{oderda2015economic,
  title={Economic burden of prescription opioid misuse and abuse: a systematic review},
  author={Oderda, Gary M and Lake, Joanita and R{\"u}dell, Katja and Roland, Carl L and Masters, Elizabeth T},
  journal={Journal of pain \& palliative care pharmacotherapy},
  volume={29},
  number={4},
  pages={388--400},
  year={2015},
  publisher={Taylor \& Francis}
}

@article{swedo2020adolescent,
  title={Adolescent opioid misuse attributable to adverse childhood experiences},
  author={Swedo, Elizabeth A and Sumner, Steven A and de Fijter, Sietske and Werhan, Luke and Norris, Kirkland and Beauregard, Jennifer L and Montgomery, Martha P and Rose, Erica B and Hillis, Susan D and Massetti, Greta M},
  journal={The Journal of Pediatrics},
  volume={224},
  pages={102--109},
  year={2020},
  publisher={Elsevier}
}

@inproceedings{Luclinicalbert,
author = {Lu, Qiuhao and Dou, Dejing and Nguyen, Thien},
year = {2022},
title = {ClinicalT5: A Generative Language Model for Clinical Text},
booktitle = "EMNLP"
}

@STRING{CVPR = "Proc. IEEE Conf. Computer Vision and Pattern Recognition"}

@STRING{ICCV = "Proc. Int'l Conf. Computer Vision"}

@STRING{ICML = "Proc. Int'l Conf. Machine Learning"}

@STRING{AAAI = "Proc. AAAI Conf. Artificial Intelligence"}

@STRING{SIGKDD = "Proc. ACM SIGKDD Int'l Conf. on Knowledge Discovery and Data Mining"}

@article{liu2023semantic,
  title={Semantic matching in machine reading comprehension: An empirical study},
  author={Liu, Qian and Mao, Rui and Geng, Xiubo and Cambria, Erik},
  journal={Information Processing \& Management},
  volume={60},
  number={2},
  pages={103145},
  year={2023},
  publisher={Elsevier}
}

@article{singhal2022large,
  title={Large Language Models Encode Clinical Knowledge},
  author={Singhal, Karan and Azizi, Shekoofeh and Tu, Tao and Mahdavi, S Sara and Wei, Jason and Chung, Hyung Won and Scales, Nathan and Tanwani, Ajay and Cole-Lewis, Heather and Pfohl, Stephen and others},
  journal={arXiv preprint arXiv:2212.13138},
  year={2022}
}

@article{singhal2023towards,
  title={Towards Expert-Level Medical Question Answering with Large Language Models},
  author={Singhal, Karan and Tu, Tao and Gottweis, Juraj and Sayres, Rory and Wulczyn, Ellery and Hou, Le and Clark, Kevin and Pfohl, Stephen and Cole-Lewis, Heather and Neal, Darlene and others},
  journal={arXiv preprint arXiv:2305.09617},
  year={2023}
}

@article{lee2020biobert,
  title={BioBERT: a pre-trained biomedical language representation model for biomedical text mining},
  author={Lee, Jinhyuk and Yoon, Wonjin and Kim, Sungdong and Kim, Donghyeon and Kim, Sunkyu and So, Chan Ho and Kang, Jaewoo},
  journal={Bioinformatics},
  volume={36},
  number={4},
  pages={1234--1240},
  year={2020},
  publisher={Oxford University Press}
}

@misc{healthcare_report,
      title={Health Online 2013}, 
      author={Susannah Fox and Maeve Duggan},
      year={2012},
      journal={Pew Research Internet Project Report},
      volume={01},
}

@inproceedings{pergola-etal-2021-boosting,
    title = "Boosting Low-Resource Biomedical {QA} via Entity-Aware Masking Strategies",
    author = "Pergola, Gabriele  and
      Kochkina, Elena  and
      Gui, Lin  and
      Liakata, Maria  and
      He, Yulan",
    booktitle = "Proceedings of the 16th Conference of the European Chapter of the Association for Computational Linguistics: Main Volume",
    month = apr,
    year = "2021",
    address = "Online",
    publisher = "Association for Computational Linguistics",
    url = "https://aclanthology.org/2021.eacl-main.169",
    doi = "10.18653/v1/2021.eacl-main.169",
    pages = "1977--1985",
}

@inproceedings{jin2019pubmedqa,
  title={PubMedQA: A Dataset for Biomedical Research Question Answering},
  author={Jin, Qiao and Dhingra, Bhuwan and Liu, Zhengping and Cohen, William and Lu, Xinghua},
  booktitle={Proceedings of the 2019 Conference on Empirical Methods in Natural Language Processing and the 9th International Joint Conference on Natural Language Processing (EMNLP-IJCNLP)},
  pages={2567--2577},
  year={2019}
}

@inproceedings{pal2022medmcqa,
  title={Medmcqa: A large-scale multi-subject multi-choice dataset for medical domain question answering},
  author={Pal, Ankit and Umapathi, Logesh Kumar and Sankarasubbu, Malaikannan},
  booktitle={Conference on health, inference, and learning},
  pages={248--260},
  year={2022},
  organization={PMLR}
}

@misc{wang2024_healthcaremagic100ken,
  author = {Wang, Rongsheng},
  title = {HealthCareMagic-100k-en},
  year = {2024},
  howpublished = {\url{https://huggingface.co/datasets/wangrongsheng/HealthCareMagic-100k-en}},
  note = {Accessed: 2024-11-13}
}

@article{he2020meddialog,
  title={Meddialog: Two large-scale medical dialogue datasets},
  author={He, Xuehai and Chen, Shu and Ju, Zeqian and Dong, Xiangyu and Fang, Hongchao and Wang, Sicheng and Yang, Yue and Zeng, Jiaqi and Zhang, Ruisi and Zhang, Ruoyu and others},
  journal={arXiv preprint arXiv:2004.03329},
  year={2020}
}

@inproceedings{banerjee2005meteor,
  title={METEOR: An automatic metric for MT evaluation with improved correlation with human judgments},
  author={Banerjee, Satanjeev and Lavie, Alon},
  booktitle={Proceedings of the acl workshop on intrinsic and extrinsic evaluation measures for machine translation and/or summarization},
  pages={65--72},
  year={2005}
}

@InProceedings{shen2023deepmad,
    author    = {Shen, Xuan and Wang, Yaohua and Lin, Ming and Huang, Yilun and Tang, Hao and Sun, Xiuyu and Wang, Yanzhi},
    title     = {DeepMAD: Mathematical Architecture Design for Deep Convolutional Neural Network},
    booktitle = {Proceedings of the IEEE/CVF Conference on Computer Vision and Pattern Recognition (CVPR)},
    month     = {June},
    year      = {2023},
    pages     = {6163-6173}
}

@article{shen2024agile, 
title={Agile-Quant: Activation-Guided Quantization for Faster Inference of LLMs on the Edge}, volume={38}, url={https://ojs.aaai.org/index.php/AAAI/article/view/29860}, DOI={10.1609/aaai.v38i17.29860}, number={17}, journal={Proceedings of the AAAI Conference on Artificial Intelligence}, author={Shen, Xuan and Dong, Peiyan and Lu, Lei and Kong, Zhenglun and Li, Zhengang and Lin, Ming and Wu, Chao and Wang, Yanzhi}, year={2024}, month={Mar.}, pages={18944-18951} }

@article{shen2025heima,
  title={Efficient reasoning with hidden thinking},
  author={Shen, Xuan and Wang, Yizhou and Shi, Xiangxi and Wang, Yanzhi and Zhao, Pu and Gu, Jiuxiang},
  journal={arXiv preprint arXiv:2501.19201},
  year={2025}
}

@article{shen2025lazydit, title={LazyDiT: Lazy Learning for the Acceleration of Diffusion Transformers}, volume={39}, url={https://ojs.aaai.org/index.php/AAAI/article/view/34248}, DOI={10.1609/aaai.v39i19.34248}, number={19}, journal={Proceedings of the AAAI Conference on Artificial Intelligence}, author={Shen, Xuan and Song, Zhao and Zhou, Yufa and Chen, Bo and Li, Yanyu and Gong, Yifan and Zhang, Kai and Tan, Hao and Kuen, Jason and Ding, Henghui and Shu, Zhihao and Niu, Wei and Zhao, Pu and Wang, Yanzhi and Gu, Jiuxiang}, year={2025}, month={Apr.}, pages={20409-20417} }

@article{shen2024edgeqat,
  title={Edgeqat: Entropy and distribution guided quantization-aware training for the acceleration of lightweight llms on the edge},
  author={Shen, Xuan and Kong, Zhenglun and Yang, Changdi and Han, Zhaoyang and Lu, Lei and Dong, Peiyan and Lyu, Cheng and Li, Chih-hsiang and Guo, Xuehang and Shu, Zhihao and others},
  journal={arXiv preprint arXiv:2402.10787},
  year={2024}
}

@article{shen2025numerical, title={Numerical Pruning for Efficient Autoregressive Models}, volume={39}, url={https://ojs.aaai.org/index.php/AAAI/article/view/34249}, DOI={10.1609/aaai.v39i19.34249}, number={19}, journal={Proceedings of the AAAI Conference on Artificial Intelligence}, author={Shen, Xuan and Song, Zhao and Zhou, Yufa and Chen, Bo and Liu, Jing and Zhang, Ruiyi and Rossi, Ryan A. and Tan, Hao and Yu, Tong and Chen, Xiang and Zhou, Yufan and Sun, Tong and Zhao, Pu and Wang, Yanzhi and Gu, Jiuxiang}, year={2025}, month={Apr.}, pages={20418-20426} }

@inproceedings{shen2024searchllm,
 author = {Shen, Xuan and Zhao, Pu and Gong, Yifan and Kong, Zhenglun and Zhan, Zheng and Wu, Yushu and Lin, Ming and Wu, Chao and Lin, Xue and Wang, Yanzhi},
 booktitle = {Advances in Neural Information Processing Systems},
 editor = {A. Globerson and L. Mackey and D. Belgrave and A. Fan and U. Paquet and J. Tomczak and C. Zhang},
 pages = {139294--139315},
 publisher = {Curran Associates, Inc.},
 title = {Search for Efficient Large Language Models},
 url = {https://proceedings.neurips.cc/paper_files/paper/2024/file/fb64a43508e0cfe53ee6179ff31ea900-Paper-Conference.pdf},
 volume = {37},
 year = {2024}
}

@article{shen2022vitlth,
  title={Data level lottery ticket hypothesis for vision transformers},
  author={Shen, Xuan and Kong, Zhenglun and Qin, Minghai and Dong, Peiyan and Yuan, Geng and Meng, Xin and Tang, Hao and Ma, Xiaolong and Wang, Yanzhi},
  journal={arXiv preprint arXiv:2211.01484},
  year={2022}
}

@ARTICLE{shen2024hotaq,
  author={Shen, Xuan and Han, Zhaoyang and Lu, Lei and Kong, Zhenglun and Dong, Peiyan and Li, Zhengang and Xie, Yanyue and Wu, Chao and Leeser, Miriam and Zhao, Pu and Lin, Xue and Wang, Yanzhi},
  journal={IEEE Transactions on Computer-Aided Design of Integrated Circuits and Systems}, 
  title={HotaQ: Hardware Oriented Token Adaptive Quantization for Large Language Models}, 
  year={2024},
  volume={},
  number={},
  pages={1-1},
  doi={10.1109/TCAD.2024.3487781}}

@inproceedings{shen2025sparse,
title={Sparse Learning for State Space Models on Mobile},
author={Xuan Shen and Hangyu Zheng and Yifan Gong and Zhenglun Kong and Changdi Yang and Zheng Zhan and Yushu Wu and Xue Lin and Yanzhi Wang and Pu Zhao and Wei Niu},
booktitle={The Thirteenth International Conference on Learning Representations},
year={2025},
url={https://openreview.net/forum?id=t8KLjiFNwn}
}

@InProceedings{shen2025quartdepth,
    author    = {Shen, Xuan and Ma, Weize and Liu, Jing and Yang, Changdi and Ding, Rui and Wang, Quanyi and Ding, Henghui and Niu, Wei and Wang, Yanzhi and Zhao, Pu and Lin, Jun and Gu, Jiuxiang},
    title     = {QuartDepth: Post-Training Quantization for Real-Time Depth Estimation on the Edge},
    booktitle = {Proceedings of the IEEE/CVF Conference on Computer Vision and Pattern Recognition (CVPR)},
    month     = {June},
    year      = {2025},
    pages     = {11448-11460}
}

@article{shen2025fastcar,
  title={Fastcar: Cache attentive replay for fast auto-regressive video generation on the edge},
  author={Shen, Xuan and Ma, Weize and Zhou, Yufa and Tang, Enhao and Xie, Yanyue and Li, Zhengang and Gong, Yifan and Wang, Quanyi and Ding, Henghui and Wang, Yiwei and others},
  journal={arXiv preprint arXiv:2505.14709},
  year={2025}
}

@article{shen2025draftattention,
  title={DraftAttention: Fast Video Diffusion via Low-Resolution Attention Guidance},
  author={Shen, Xuan and Han, Chenxia and Zhou, Yufa and Xie, Yanyue and Gong, Yifan and Wang, Quanyi and Wang, Yiwei and Wang, Yanzhi and Zhao, Pu and Gu, Jiuxiang},
  journal={arXiv preprint arXiv:2505.14708},
  year={2025}
}

@INPROCEEDINGS{liu2025rora,
  author={Liu, Jun and Kong, Zhenglun and Dong, Peiyan and Shen, Xuan and Zhao, Pu and Tang, Hao and Yuan, Geng and Niu, Wei and Zhang, Wenbin and Lin, Xue and Huang, Dong and Wang, Yanzhi},
  booktitle={ICASSP 2025 - 2025 IEEE International Conference on Acoustics, Speech and Signal Processing (ICASSP)}, 
  title={RoRA: Efficient Fine-Tuning of LLM with Reliability Optimization for Rank Adaptation}, 
  year={2025},
  volume={},
  number={},
  pages={1-5},
  doi={10.1109/ICASSP49660.2025.10889613}}

@article{liu2025toward, title={Toward Adaptive Large Language Models Structured Pruning via Hybrid-grained Weight Importance Assessment}, volume={39}, url={https://ojs.aaai.org/index.php/AAAI/article/view/34078}, DOI={10.1609/aaai.v39i18.34078}, number={18}, journal={Proceedings of the AAAI Conference on Artificial Intelligence}, author={Liu, Jun and Kong, Zhenglun and Zhao, Pu and Yang, Changdi and Shen, Xuan and Tang, Hao and Yuan, Geng and Niu, Wei and Zhang, Wenbin and Lin, Xue and Huang, Dong and Wang, Yanzhi}, year={2025}, month={Apr.}, pages={18879-18887} }

\clearpage
\newpage

\clearpage
\newpage

\appendix
\section*{Appendix}

\section{QA Generation Algorithm}

\begin{algorithm}[h]
\small{
    \caption{QA Data Generation Process}
    \label{alg:qa_generation_simple}  
    \SetKwInOut{Input}{Input}  
    \SetKwInOut{Output}{Output}  
    \Input{  
        Document set $\mathcal{D}$; \\
        Persona pool $\mathcal{P}$; \\
        Desired number of QA pairs per document $N_{\text{QA}}$; \\
        Maximum attempts per document $M$  
    }  
    \Output{QA dataset $\mathcal{QA}$}  
  
    Initialize $\mathcal{QA} \leftarrow \emptyset$\;  
  
    \ForEach{document $D \in \mathcal{D}$}{  
        Initialize QA count $n \leftarrow 0$, attempt count $m \leftarrow 0$\;  
        \While{$n < N_{\text{QA}}$ \textbf{and} $m < M$}{  
            Sample personas $\mathcal{P}_D$ from $\mathcal{P}$ based on CLIP tags\;  
            \ForEach{persona $P \in \mathcal{P}_D$}{  
                Create prompt by combining document $D$ and persona $P$\;  
                Generate question $q$ using GPT-4o\;  
                \If{$q$ is answerable}{  
                    Generate answer $a$ with referred page using GPT-4o\;
                    \If{$(q, a)$ pair is decomposable}{
                        Decompose $(q, a)$ into QA sequence $\{(q_i, a_i)\}_{i=1}^{s}$\;
                        \If{$s>1$}{
                            Add $\{(q_i, a_i)\}_{i=1}^{s}$ to $\mathcal{QA}$\;
                        }
                    } 
                    $n \leftarrow n + 1$\;  
                }  
            }  
            $m \leftarrow m + 1$\;  
        }  
    } 
}
\end{algorithm}

\section{Comparison to GPT-4}

\begin{figure*}[h]
\centering
\includegraphics[width=\textwidth]{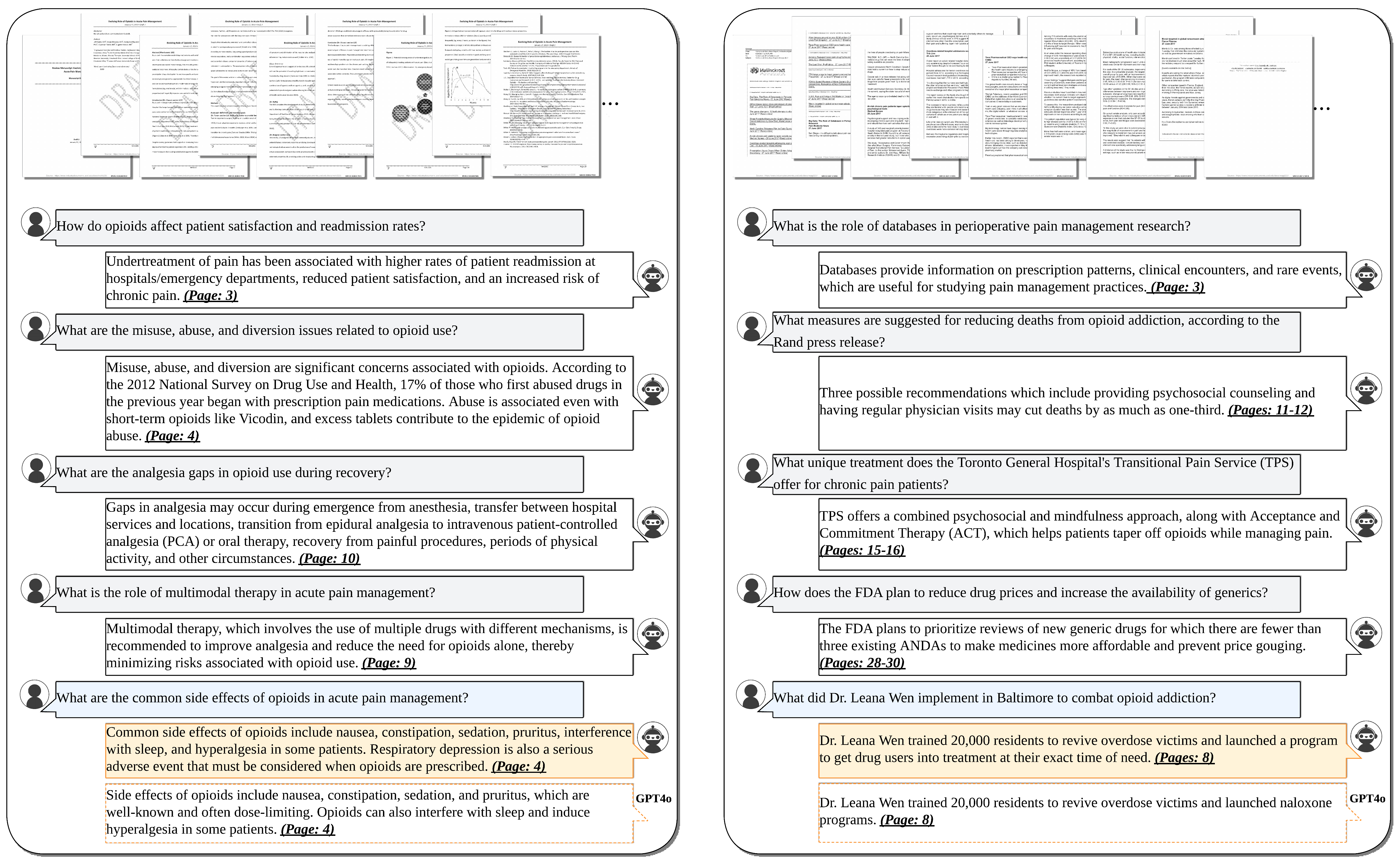}
\caption{Comparisons between our multi-page document QA model and GPT-4 on two examples. The first example is a 19-page document, and the second is a 35-page document. Due to space limitations, we only visualize a portion of the results. Our model demonstrates results comparable to GPT-4, with accurate page grounding that correctly matches the document's pages.}
\label{fig:oida_vis_samples}
\end{figure*}

\end{document}